\documentclass[11pt]{article}

\usepackage[margin=1in]{geometry}
\usepackage{fancyhdr}
\usepackage{amssymb}
\usepackage{amsmath}
\usepackage{centernot}	
\usepackage{hyperref}
\usepackage{mathrsfs}
\usepackage{mathtools}
\usepackage{txfonts}
\usepackage{tikz}
\usepackage{xcolor}
\usepackage{algorithm}
\usepackage{algorithmicx}
\usepackage{algpseudocode}
\usepackage{centernot}
\usepackage[mathscr]{euscript}
\usepackage{etoolbox}
\newtheorem{definition}{Definition}
\newtheorem{theorem}{Theorem}
\newtheorem{lemma}{Lemma}
\newtheorem{conjecture}{Conjecture}
\newtheorem{corollary}{Corollary}
\newtheorem{proof}{Proof}
\AtBeginEnvironment{quote}{\singlespacing\small}

\title{Improving Accuracy of Permutation DAG Search using Best Order Score Search\footnote{This draft will be replaced eventually by a corrected and more complete draft; please don't quote from it. The author wishes to thank Kun Zhang, Clark Glymour, Peter Spirtes and Wayne Lam for helpful discussions and suggestions, though errors and omissions (of which there are surely many) are all due to the author.}}
\author{Joseph Ramsey\\
Carnegie Mellon University\\ jdramsey@andrew.cmu.edu\\
}
\date{\today}

\begin{document}
\pagestyle{fancy}
\fancyhead{}
\fancyhead[L]{\textbf{Best Order Score Search}}
\fancyhead[R]{\textbf{Joseph Ramsey, jdramsey@andrew.cmu.edu}}
\renewcommand{\footrulewidth}{1pt}

\newcommand{\CI}[0]{\perp\!\!\!\perp}
\newcommand{\nCI}[0]{\centernot{\CI}}
\newcommand{\ot}[0]{\leftarrow}
\newcommand{\tot}{\leftrightarrow}
\newcommand{\oo}[0]{\multimapboth}
\newcommand{\oright}{\,\mathrlap{\ot}\,\,{\multimap}\,}
\newcommand{\oleft}{\,\mathrlap{\multimapinv}\,\to}

\maketitle

\begin{abstract}
\noindent The Sparsest Permutation (SP) algorithm is accurate but limited to about 9 variables in practice; the Greedy Sparest Permutation (GSP)  algorithm is faster but less weak theoretically. A compromise can be given, the Best Order Score Search, which gives results as accurate as SP but for much larger and denser graphs. BOSS (Best Order Score Search) is more accurate for two reason: (a) It assumes the ``brute faithfuness'' assumption, which is weaker than faithfulness, and (b) it uses a different traversal of permutations than the depth first traversal used by GSP, obtained by taking each variable in turn and moving it to the position in the permutation that optimizes the model score. Results are given comparing BOSS to several related papers in the literature in terms of performance, for linear, Gaussian data. In all cases, with the proper parameter settings, accuracy of BOSS is lifted considerably with respect to competing approaches. In configurations tested, models with 60 variables are feasible with large samples out to about an average degree of 12 in reasonable time, with near-perfect accuracy, and sparse models with an average degree of 4 are feasible out to about 300 variables on a laptop, again with near-perfect accuracy. Mixed continuous discrete and all-discrete datasets were also tested. The mixed data analysis showed advantage for BOSS over GES more apparent at higher depths with the same score; the discrete data analysis showed a very small advantage for BOSS over GES with the same score, perhaps not enough to prefer it.
\end{abstract}
`
\section{Overview}

Searching for causal models is a well-established activity by this point, though it's not without detractors. For one thing, it's thought by many that the only correct method for establishing causal relationships is through intervention, as is the case with a randomized control trial or when intervening on the activity of particular genes. Doing the inference from observational data alone can from this perspective be seen as second-rate. This is not completely unreasonable. A genuine reason one might think observational inference is second-rate, for instance, is that many causal search algorithms, even under (true) assumptions of linearity and Gaussianity, say, can have performance statistics that are on the low side, so that their inferences cannot be relied upon. This is related to another reason. One might think that theoretical assumptions for an algorithm are too strong for the sorts of cases that are studied. For instance, there may be latent variables which the algorithm assumes away, or there may be cycles in the true model where the algorithm assumes acyclicity. But the basic case has for a long time been causally sufficient acyclic search, such as is done using the PC (\cite{spirtes2000causation}) algorithm or the GES (\cite{chickering2002optimal})) algorithm, or the GSP algorithm (\cite{solus2017consistency}), among many others. This is not without its problems either, whether it be in terms of performance or in terms of strength of assumptions. Let's focus on this case and consider strength of assumptions. One way to make assumptions weaker is to use the SP algorithm (\cite{raskutti2018learning}), which Raskutti and Uhler show relies on a weaker assumption than the faithfulness that is assumed by PC or some weaker assumption (unclear) that is assumed by GES. SP can be very accurate, sensibly interpreted, though it is limited to a maximum of about 9 variables. Is there an algorithm that's equally accurate in most cases but can scale to larger problems without losing accuracy?

The SP algorithm itself gives a clue. This is an algorithm that, like the one introduced in Teyssier and Kohler (\cite{teyssier2012ordering}), proceeds through pure enumeration of permutations. For each such permutation one builds a DAG using a procedure given by Verma and Pearl (\cite{verma1990equivalence}), where for parents of Y in the permutation, on uses variables X such that $Y \perp\!\!\!\perp X | \{ prefix(X, O) \setminus \{X\} \}$, where O is the causal order, and $prefix(X, O)$ selects the nodes previous to $X$ in $O$. If there are $n$ variables, then for each of the possible $n!$ permutations one builds a DAG in this way and records the number of edges. One then selects the permutation that implies the graph with the smallest number of edges. Raskutti and Uhler prove that the equivalence class of this DAG is unique and correct under the assumption of faithfulness and is also unique and correct in many other situations, so that assumption that an algorithm gets a DAG in the correct Markov equivalence class (MEC) for the same cases as SP weaker than faithfulness.

The problem with SP is that it's very computationally complex, since all permutations must be visited, and that problem is super-exponential. The advice that Solus et al. give (advice that Teyssier et al. had previously given) is to visit permutations in a depth-first way that greedily optimizes some score. They suggest two scores; one is edge count, the other BIC. These scores may both readily be used. Teyssier et al. give a clever scoring procedure which can used directly, pointing out that when two adjacent variables in a permutation are swapped, one only needs to re-score those two variables, the rest of the variables all receiving the same (variable) score as before the swap. The algorithm that Teyssier et al. give is to traverse permutations in a depth first order, with a greedy hill-climbing procedure with a TABU list as a meta-loop. They report that scores increase with this procedure but do not compare result graphs back to true graphs in simulation. Solus et al. also do a depth-first traversal of permutation space with their GSP algorithm, making use of Chickering's (\cite{chickering2002optimal}) theory of covered edge reversal. Judging from the implementation of GSP in the Python package causaldag (\cite{solus2017consistency}, \cite{lu2021improving}), this procedure is fast but has difficulty scaling accurately to moderate or large sized graphs.

Since SP works so well for small models, perhaps a different way of traversing the space of permutations might be explored that gets closer to the theoretical guarantees in Raskutti and Uhler. Below are some suggestions. But first the topic of path cancellation needs to be generalized to a condition that applies not just to linear models but more generally.

\section{Approach}

\subsection{Preliminaries}

To situate the discussion, the following definitions are given. A \textit{dataset} is a measurement of a set of variables for each of several different i.i.d. points. he dataset is assumed to be i.i.d. A \textit{variable} is a named distribution of a single column in such a dataset.Also, for the base case linear relationships between the variables will be assumed, with Gaussian disturbances. A \textit{directed edge} is of the form $X \rightarrow Y$ for variables $X$ and $Y$. A \textit{directed path} takes the form $X \rightarrow .... \rightarrow Y$ from some $X$ to some $Y$. A \textit{cycle} is a directed path from a variable to itself. A \textit{directed graph} is a pair $<V, E>$ where $V$ is a set of variables and $E$ is a set of directed edges over $V$. A directed graph is \textit{acyclic} if it contains no cyclic paths. $X$ is \textit{d-separated} from $Y$ conditional on a set $Z$ just in case that statistical relationship holds; otherwise, it is \textit{d-connected}. \textit{D-separation}, by contrast, is a \textit{graphical} relationship, where $X$ is \textit{d-separated} from $Y$ conditional on $Z$ just in case every collider on a path from $X$ to $Y$ is either in $Z$ or does not contain a node in the descendants of any variables in $Z$. Otherwise, the variables are \textit{d-connected}. The Markov and faithfulness assumptions are defined as follows. Here (and below) $G^*$ is a graph and $P$ is a probability distribution over the variables of $G^*$; call $(G^*, P)$ a \textit{model}.

\begin{definition}
Model $(G^*, P)$ satisfies the Markov assumption if $dsep_{G^*}(X, Y | Z)$ implies $X \perp\!\!\!\perp_{P} Y | Z$.
\end{definition}

\begin{definition}
Model $(G^*, P)$ satisfies the faithfulness assumption if $X \perp\!\!\!\perp_{P} Y | Z$ implies $dsep_{G^*}(X, Y | Z).$
\end{definition}

\noindent For a linear model,two paths from $X$ into $Y$ \textit{cancel} if the correlation of $X$ and $Y$ is zero; in this case, an extra conditional independence is added to the model. A \textit{permutation} $O$ of a set of variables $V$ is a linear ordering of $V$. A \textit{causal order} of a graph $G^*$ is a permutation $O$ such that each ancestor of each variable occurs before that variable in $O$. The \textit{prefix} of a variable $X$ in a permutation $O$ is the set of variables that occur in $O$ at earlier indices than $X$. A \textit{measured variable} is a variable in a dataset for which data are given row-wise. A \textit{latent variable} is a variable that may affect the measured variables in a dataset causally but which is not in the dataset. It is assumed that there are no latent variables and no cycles. A \textit{Markov blanket} of a variable $X$ with respect to a set $X \in S = mb(X, S)$ is a minimal set such that if $Y \neq X$ and $Y \not \in M$ but $Y \in S$, then $Y \perp\!\!\!\perp X | S$. Precision and recall are given the usual definitions--that is, if $TP$ is the number of true positives, $FP$ the number of false positives, and $FN$ the number of false negatives, then Precision = $TP / (TP + FP)$ and Recall = $TP / (TP + FN)$. Structural Hamming Distances as in \cite{tsamardinos2006max}. In graph G, $x$ is a parent of $y$ just in case $x \rightarrow y$. Two variables x and y are adjacent in graph G just in case $x \rightarrow y$ or $y \rightarrow x$. A \textit{triangle} is a triplet of variables $<x, y, z>$ such that x and y are adjacent, y and z are adjacent, and x and z are adjacent. 

\subsection{Path Cancellation}

Consider path cancellation, a type of unfaithfulness that affects linear models, for instance. Here, something that has perhaps not been clear, Verma and Pearl's \cite{verma1990equivalence} method for building DAGs handles path cancellation just fine. This is Pearl's method:

\begin{definition}
\label{Pearl}
A DAG is built by Verma and Pearl's method by finding for each variable $X$ the set of variables $Y$ such that $X \perp\!\!\!\!/\!\!\!\!\!\perp Y | (prefix(X, O) \setminus \{Y\})$, and making those variables parents of $X$.
\end{definition}

The reason this addresses the path cancellation issue is that, given a true causal order, all prefix variables (other than a putative parent) are conditioned on when building a DAG using Verma and Pearl's method. As an example, consider a 4-node path-canceling example with the graph $1 \rightarrow 4, 1 \rightarrow 2 \rightarrow 3 \rightarrow 4$, parameterized as a standardized linear, Gaussian model, and imagine that the two paths $1 \rightarrow 4$ and $1 \rightarrow 2 \rightarrow 3 \rightarrow 4$ cancel, creating an unfaithful Independence $1 \perp\!\!\!\perp 4)$. These effect of the $1 \rightarrow 4$ edges may be \textit{cancelled}, in the sense that this unfaithful independence may be formed as a result of careful choice of coefficients for a standardized model, so that total effect of 1 on 4 is zero. This can be accomplished by assigning the coefficient for the $1 \rightarrow 4$ edge the coefficient $-0.125$ and the coefficients for the other three edges all $0.5$, so that $1 \perp\!\!\!\perp 4)$. Say one wants to know what the parents of 4 are. The PC algorithm (\cite{spirtes2000causation}) would remove 1 as a parent of 4 in the first round or processing, because of the unconditional independence of 1 and 4. The GES algorithm (\cite{chickering2002optimal}), by contrast, through a bit of reasoning, would eventually add 3 to the conditioning set for 4 and conclude (with this condition) that 1 and 4 are dependent, and therefore add 1 as a parent of 4. A permutation algorithm using Pearl's method for building a DAG given a true causal order ($<1, 2, 3, 4>$ in this case) agrees with GES on this point, in spirit, since the question for it will be whether $1 \perp\!\!\!\!/\!\!\!\!\!\perp 4 | 2, 3$, which includes conditioning on 3, thus blocking the second path and ``breaking'' the cancellation. So a conditional dependence will be found. This reasoning generalizes, in the following sense. If $X \rightarrow Y$ and there are several paths into $Y$ such that the effect of the $X \rightarrow Y$ edge is canceled, Verma and Pearl's algorithm will nevertheless find the $X \rightarrow Y$ edge, given a true causal order, by conditioning on all of the parents of $Y$ other than $X$, thus blocking all paths into $Y$ other than the path $X \rightarrow Y$ itself.

Note that the situation of allowing an effect of a parent to be cancelled through careful selection of coefficients gives a correlate faithfulness assumption that is more general than unfaithfulness itself. In the model above, for instance, one can't have $2 \perp\!\!\!\perp 3$ with an additional edge $2 \rightarrow 3$ in the model--that would be an example of unfaithfulness \textit{not} due to a path cancellation. Let $(G^*, P)$ be a model, where the model is linear. Let $(G^*, P)$ satisfy \textit{brute faithfulness} just in case $(G^*, P)$ satisfies the faithfulness assumption, except for examples of path cancellation. Thus, a model may satisfy the brute faithfulness assumption even it does not satisfy the faithfulness assumption, provided all spurious independencies in the model are due to path cancellation.

A problem with Verma and Pearl's method is purely algorithmic; by conditioning on all variables in the prefix of $Y$ other than $X$, for large models, the number of variables conditioned on may become sizeable, slowing the independence test down and potentially making it less accurate. To condition on smaller sets, an iterated Grow-Shrink method may be employed as given in Algorithm \ref{GrowShrink}. The latter is a simple adaptation of the Grow-Shrink algorithm (\cite{margaritis1999bayesian}). The problem with path cancellation, with Grow-Shrink, is that some parents, say, of a variable $Y$ may not be included in the Markov blanket on the first pass, due to path cancellation, but if one conditions on a proper subset of variables that are included, a conditional dependence may be found for these parents, and then, on a second pass, they will be included. Thus, the Grow-Shrink algorithm, iterated, will eventually find the variables to condition on, and when it does find the conditioning, the dependence will be found (e.g., in the above example, conditioning on $3$). Eventually all parents, children, or parents of children will be included that would have been included if the stronger condition, faithfulness, held instead. That is,

\begin{lemma}
Let $G^*, P$ be a model. If there is a set $S \subseteq parents(X, G)$ such that $X \perp\!\!\!\!/\!\!\!\!\!\perp Y | S)$, Algorithm \ref{GrowShrink} will find one.
\end{lemma}

\noindent Algorithm \ref{dag} may use either Verma and Pearl's method or the iterated Grow-Shrink method of Algorithm \ref{GrowShrink}. Note that the iterated Grow-Shrink algorithm is not \textit{equivalent} to the Verma-Pearl algorithm, since as in the above example both 2 and 3 will be conditioned on, whereas for the iterated Grow-Shrink algorithm this dual conditioning will never be done, but it does have the same property of breaking path cancellation.

If one's model is not linear, ``path cancellation'' may seem ill-fit as a description, even though the same situation may be faced of extra independencies being added to the underlying distribution over and above what one might expect assuming faithfulness. Here is a definition to clarify what brute faithfulness corresponds to in terms of d-separation and independence.

\begin{definition}
Model $(G^*, P)$ satisfies the generalized brute faithfulness assumption if for each causal order $O$ of $G^*$ and for each parent $Y$ of $X$ in $G^*$ there exists a subset $Z \subseteq prefix(X, O) \setminus \{Y\}$ such that $X \perp\!\!\!\!/\!\!\!\!\!\perp Y | Z)$.
\end{definition}

\noindent Models that satisfy brute faithfulness also satisfy generalized brute faithfulness, though the latter allows for the possibility of analyzing nonlinear data. In terms of linearity, this exercise is meant to find effective conditional independencies for parents that are ineffective due to path cancellation, as in the simple 4-variable path canceling example above. The method is implemented by the iterated Grow-Shrink algorithm (Algorithm \ref{GrowShrink}). Note that recursive linear models satisfy the brute faithfulness assumption:

\begin{corollary}
Let $(G^*, P)$ be a recursive linear model. Then $(G^*, P)$ satisfies the brute faithfulness assumption.
\end{corollary}

\begin{proof}
This follows from the observation that the only way an ``extra'' conditional independencies may arise in such a model (that is, not through finite sample effects) is through path cancellation. All other conditional dependencies and independencies are implied by the Markov and faithfulness assumptions themselves.
\end{proof}

\noindent Other types of models more generally may also satisfy the brute faithfulness assumption; it would have to be shown for them of course, but the current theory may still be very helpful.

It follows that for a permutation algorithm, for the linear case, so long as a correct permutation is eventually found, adjacency recall will be $100\%$. This suggests the following lemma.

\begin{lemma}
\label{BOSS-brute}
If $M = mb(x, O) \subseteq prefix(x, O)$ under the assumption of faithfulness, then under the assumption of brute faithfulness for the linear case, $mb(x, O)$ is also equal to $M$.
\end{lemma}

\begin{proof}
This follows immediately from the definition of brute faithfulness and the above Corollary.
\end{proof}

\begin{corollary}
\label{DAG-brute}
Let $O$ be a correct causal order and assume brute faithfulness for the linear case. Then Algorithm \ref{dag} will build a DAG in the correct MEC.
\end{corollary}

\begin{proof}
Immediate.
\end{proof}

\noindent To ensure correct orientations have been found under brute faithfulness, it is enough to assume a correct causal order has been found.

\subsection{A Different Kind of Permutation Traversal}
\label{PermutationTraversal}

The next topic is permutation traversal. Teyssier and Kohler (\cite{teyssier2012ordering}) and Solus et al. (\cite{solus2017consistency}), following Teyssier and Kohler, pursue a depth-first traversal of permutations. But there is another, more breadth-first, way of traversing permutations that addressed the accuracy problem mentioned above and, combined with the assumption of brute unfaithfulness, gives results on par with those that would be given by SP, even for large or fairly dense graphs, for much higher average degrees for the true graph. 

The method is simple. One starts with a random permutation, or whatever order of variables is given in the data. (No careful selection of initial permutations is needed.) First one takes a variable and relocates it to every other position in the permutation (Algorithm \ref{BOSS}). For each order adjustment, a DAG is built (Algorithm \ref{dag}), either using Pearl's algorithm or the faster iterated Grow-Shrink algorithm (Algorithm \ref{GrowShrink}). A score is recorded for each DAG. The permutation is then chosen with the optimal score, in effect moving the variable to an optimal position for score. Next, one takes up the second variable and does the same thing. One continues in this way until all variables have been processed, and then one goes back to the beginning of the list and processes the variables again. One repeats this whole procedure until no more changes can be made in this way. A simple worked example is given in Section \ref{SimpleWorkedExample}.

Sometimes one can reach a final permutation using this methods that does not have the optimal score. The strategy in this case is to recognize that this must be a example where moving two variables simultaneously in the permutation will increase the score. One should look for forks to convert to colliders. A good strategy for this, for a variable $X$ in $O$ is to consider each distinct pair of adjacent variables $Y$ and $Z$ to the right of $X$ in $O$, with $Y*-*Z$ and then reverse the $X*-*Y$ edge, then reverse the $X*-*Z$ edge, so that now $Y$ and $Z$ are to the left of $X$ but the built graph is otherwise unperturbed. If this results in the shield $Y*-*Z$ being removed from the built graph, the score will decrease. Otherwise, the variables are restored to their original positions and the search continued. This is what Algorithm \ref{TwoStepCorollary} does.

Let $score_{edge}(O)$ be as defined in the pseudocode below. The main result needed is the following:

\begin{theorem}
\label{RelocationTheorem}
Let model $(G^*, P)$ satisfy faithfulness or brute faithfulness and let O be a permutation over the variables of $G^*$. Then if $O$ is a permutation that is not a causal order of $G^*$, then there exists a variable $X$ in $O$, which when relocated, or a pair of variables $X$ and $Y$ in $O$ which, when simultaneously relocated, will yield a permutation $O'$ that does not make $score_{edge}(O)$ worse.
\end{theorem}

\begin{proof}
Assuming faithfulness or brute faithfulness, DAGs will be built correctly in Verma and Pearl's sense. We proceed by induction. We have by faithfulness, say, that all true adjacencies correspond to dependencies in the graph, so the question is how many extra dependencies there are. These extra dependencies will correspond to shields added for unshielded triples that are not oriented correctly by the permutation. Consider the first variable $X$. If $X$ is not exogenous, find a causally exogenous variable $Y$ at position $n$ in $O$ and put it before $X$ in the permutation, yielding permutation $O'$. This may removes shields, if children of the $Y$ occurred before $Y$ in its original position in the permutation. Shields between position $1$ and position $n$ in $O$. The ordering of the other variables in $O$ remains unchanged in $O'$, so any extra shields among other variables for $O$ are still extra shields for $O'$, so $score{edge}$ will not be worse than before. Likewise, if the first $k$ variables in $O$ are in causal order and the $k + 1$th variable is not, find some variable $Z$ at position $m$, or some pair of variables $Z_1, Z_2$ at positions $m_1, m_2$, say, such that $Z$ extends the causal order, or $Z_1$, $Z_2$ extend extend the causal order and move these to position $k + 1$, or optionally as well, $k + 2$, yielding permutation $O''$. It may be necessary to move a pair of variables in order to reorient a fork as a collider otherwise, single moves will do. It is not necessary to more more than two nodes simultaneously, since the goal is only to remove shields for unshielded triples in the true DAG. The same argument applies as above, except that now $Z$, or $Z_1$ and $Z_2$, may have parents that occur (correctly) before position $k + 1$ in $O''$. Again, additional shields for variables other than $Z$, $Z_1$, or $Z_2$ will be left unchanged, so $score_{edge}$ will not be wore than before.
\end{proof}

\noindent Theorem \ref{RelocationTheorem} does not give unambiguous advice for how to construct an algorithm. The procedure BOSS adopts is to to follow all one-edge moves for a given variable, then when there are no more such moves, look for a two-step relocation of adjacent subsequent variables in the permutation that can be made. The procedure is repeated until no more improvements can be made. The two-step moves can be characterized as follows:

\begin{corollary}
\label{TwoStepCorollary}
Let $G, P$ satisfy faithfulness or brute unfaithfulness and let permutation $P_m$ be such that for all relocation permutations $P_r$ of $P_m$, $score_{edge}(P_r) \geq score_{edge}(P_m)$. Then if $score_{edge}(P_m)$ is not optimal, Algorithm \ref{TWO_STEP} can find a permutation with a lower score.
\end{corollary}

\begin{proof}
Sketch. Assuming faithfulness or brute faithfulness, DAGs will be built correctly in Verma and Pearl's sense. If no single relocation move improves the score, then one needs to consider moving two nodes at a time, by Theorem \ref{RelocationTheorem}, before concluding that no further moves can be made to reduce the score. But a simple enumeration shows that the only relevant simultaneous two-edge adjustment that cannot be accomplished by a sequence of single-edge adjustments are those that convert shielded forks to an unshielded colliders. So let $X \in P_m$ and consider pair of distinct variables $Y$ and $Z$ adjacent to $X$ in $dag(P_m)$ that are both causally prior to $X$. If $Y$ and $Z$ are both already in $prefix(X, P_m)$, then if $Y$ and $Z$ are not adjacent in that graph, nothing further needs to be done. WLOG, let $Y \in prefix(X, P_m)$ and let $Z \not \in prefix(X, P_m)$. Then $Z$ may be moved into $prefix(X, P_m)$ by a shield-eliminating single move, if $Y$ and $Z$ are not adjacent in the true graph, against the assumption that all single-edge moves that can improve the score have already been made. Thus, it must be that $Y$ and $Z$ are both not in $prefix(X, P_m)$. By iterating over all distinct adjacent variables to the right of $X$, then, one can find such a two-variable move to make to convert a shielded fork into an unshielded collider if such a move exists. This is what Algorithm \ref{TWO_STEP} does.
\end{proof}

\noindent If a two-step move can be made, one makes it and then looks for as many new one-step moves to make as possible, forming a maximal sequence of scores, then repeats the procedure until convergence. The result will be a model with the minimal number of edges.

The edge count score is used above, following Raskutti and Uhler (\cite{raskutti2018learning}). Solus et al. also suggesting using BIC as a score. One way to motivate this is to notice that in the limit of large sample BIC gets the best score (according to Schwarz (\cite{schwarz1978estimating})) for model $(G^*, P)$ where $P$ has an  elliptical distribution. The model found where the best score is attained is in the true MEC with worse scores otherwise, under faithfulness. This is true also under brute unfaithfulness where DAGs are built using either Verma and Pearl's procedure or the iterated Grown-Shrink algorithm. Notably, it does not matter which type of data the BIC score is for, so it's possible to analyze linear, Gaussian continuous data or discrete data or data with both continuous and discrete columns, all with about the same accuracy (though different speeds). This gives the BOSS algorithm a certain degree of generality with respect to variable types, which we will test.

This follows:

\begin{corollary}
\label{BOSS-faithfulness}
Assuming brute faithfulness or generalized brute faithfulness, Algorithms \ref{BOSS} will find models in the correct MEC.
\end{corollary}

\begin{proof}
Immediate.
\end{proof}

It is helpful to consider how weak an assumption BOSS can make and still return graphs in the correct MEC. One can certainly make a weaker assumption than faithfulness, since brute faithfulness suffices. This makes it theoretically more general than, say, the PC algorithm (\cite{spirtes2000causation}). Nevertheless, BOSS is not making a weaker assumption than SMR, the assumption of the SP algorithm, which judges each permutation individually and returns the sparsest. One may conjecture that BOSS is correct under the SMR assumption, since by construction, as noted above, for the linear case, under the brute faithfulness assumption, a graph with the minimal number of edges is returned. Some comments from Raskutti and Uhler--this definition:

\begin{definition}
Raskutti et al., \cite{raskutti2018learning}. A model $(G^*, P)$ satisfies the sparsest Markov representation (SMR) assumption if ($G^*, P)$ satisfies the Markov assumption and $|G| > |G^*|$ for every DAG $G$ such that $(G^*, P)$ satisfies the Markov assumption and $G \not \in M(G^*)$.
\end{definition}

\noindent Raskutti and Uhler prove from this and other propositions the following:

\begin{theorem}
Raskutti et al., \cite{raskutti2018learning}. The SP algorithm outputs $G \in M(G^*)$ if and only if the model $(G^*, P)$ satisfies the SMR assumption.
\end{theorem}

\noindent Verma and Pearl have the following idea, as Raskutti and Uhler point out: "Pearl (1988, Theorem 9 on p. 119) and also Verma and Pearl (1988) showed that for any positive measure P and any permutation $\pi$, $P$ is Markov to $G^*$ and satisfies the minimality assumption, meaning that there is no proper sub-DAG of $G^*$ that satisfies the Markov property." We may make the following parallel conjecture:

\begin{conjecture}
\label{BOSS_SMR}
The BOSS algorithm outputs $G \in M(G^*)$ if and only if the model $(G^*, P)$ satisfies the SMR assumption.
\end{conjecture}

\noindent If this is the case, even if only in the backward direction, then one expects BOSS (with the two-step procedure) to output DAGs in the same MEC as SP. Solus et al. identify a sequence of implications of assumptions and give examples to show that the implications are strong--e.g., that ESP, the assumption under which the algorithm ESP identifies a model in the true MEC, is weaker than the corresponding assumption for TSP. For this case they give an example where ESP gets the answer given by SP but TSP does not. (ESP, TSP, and GSP are the algorithms by those names defined in Solus et al. \cite{solus2017consistency}.) These counterexamples are collected up in the Section \ref{UnfaithfulExamples} and studied. Conjecture \ref{BOSS_SMR} implies that if SP returns a model, BOSS will return the same model. On repeated examination from random starting points, for each counterexample, this is the case. Note that these are not linear examples but are rather specified simply as lists of conditional independencies (by the original authors).

Pseudocode for the BOSS algorithm and related procedures follows.

\begin{algorithm}
\caption{Score-based Iterated GrowShrink} 
\label{GrowShrink}
\end{algorithm}
\begin{algorithmic}[1]
\Function{$\textit{GrowShrinkMb}$}{v, $V=<V_1,..,V_n>$}
\State $mb \gets$ an empty graph over V
\While{score improves}
    \For {$e$ in $V$}
        \State $v \gets null$
        \If {$mb$ does not contain $e$}
            \State Add $e$ to $mb$
            \State $s \gets BIC(v | mb)$
            \If {$s < s_0$}
                \State $s_0 \gets s$
                \State $v \gets e$
            \EndIf
            \State Remove $e$ from $mb$
        \EndIf
    \EndFor
    \If {$v \neq null$}
        \State Add $v$ to $mb$
    \EndIf
    \State $w \gets null$
    \For {{$e$ in $mb$}}
        \State Remove $w$ from $mb$
        \State s $\gets BIC(n | mb)$
        \If {$s \leq s_0$}
            \State $s_0 \gets s$
            \State $w \gets e$
        \EndIf
    \EndFor
    \If {$w \neq null$}
        \State Remove $w$ from $mb$
    \EndIf
\EndWhile
\State Return $mb$
\EndFunction
\end{algorithmic}

\begin{algorithm}
\caption{Score a permutation)} 
\label{Score}
\end{algorithm}
\begin{algorithmic}[1]
\Function{$score_{edge}$}{$V=<V_1,..,V_n>$}
\State $sum \gets 0$
\For {$i$ in $1:|V|)$}
    \State $sum \gets sum + \lvert GrowShrinkMb(V(i) \rvert$
\EndFor
\State Return $sum$
\EndFunction
\end{algorithmic}
 jj
\begin{algorithm}
\caption{Builds a DAG given a permutation)} 
\label{dag}
\end{algorithm}
\begin{algorithmic}[1]
\Function{$\textit{dag}$}{$b_0=<V_1,..,V_n>$}
\State {$G \gets $ empty graph over $b_0$}
\State $sum \gets 0$
\For {$v$ in $b_0$}
    \State {$mb \gets GrowShrinkMb(v, prefix(v, b_0))$}
    \For {$w$ in $mb$}
        \State {Add $w \rightarrow v$ to G}
    \EndFor
\EndFor
\State Return $sum$
\EndFunction
\end{algorithmic}

\begin{algorithm}
\caption{Best Order Score Search (BOSS)} 
\label{BOSS}
\end{algorithm}
\begin{algorithmic}[1]
\Function{$\textit{BOSS}$}{$V=<V_1,..,V_n>$}
\State $b \gets V$
\State $s \gets score_{edge}(b)$
\While{$score_{edge}$ improves}
    \While{$score_{edge}(b)$ improves}
        \For {$v$ in $V$}
            \State Move $v$ to the last position in $b$, yielding $b'$
            \State $s \gets score_{edge}(b)$
            \While {changed}
                \State Move $v$ one index to the left in $O$
                \If {$score_{edge}(b) < s$}
                    \State $b' \gets b$
                    \State $s \gets score_{edge}(b)$
                \EndIf
            \EndWhile
            \State {$b \gets b'$}
        \EndFor
    \EndWhile
    \State $b \gets twostep(b)$
\EndWhile
\State Return $b$
\EndFunction
\end{algorithmic}

\begin{algorithm}
\caption{Two-step procedure, to be applied after Algorithm \ref{BOSS} otherwise produces no more changes} 
\label{TWO_STEP}
\end{algorithm}
\begin{algorithmic}[1]
\Function{$\textit{twostep}$}{V}
\State $b \gets V$
\State $s \gets score_{edge}(b)$
\State $o \gets V$
\While{$score_{edge}$ improves}
    \State $b \gets V$
    \State $s \gets score_{edge}(b)$
    \State $v \gets o(i)$
    \For {$v$ in $O$}
        \For {$i_{r_1}$ from $index(v, b)$ to $|b|$}
            \For {$i_{r_2}$ from $i_{r_1}$ to $|b|$}
                \State $r_1 \gets b(i_{r_1})$
                \State $r_2 \gets b(i_{r_2})$
                \If {$v, r_1, r_2$ form a triangle}
                    \State $b' \gets b$
                    \State Swap $v$ and $r_1$ in $b$
                    \State Swap $v$ and $r_2$ in $b$
                    \If {($score_{edge}(b) < score_{edge}(b')$}
                        \State Return true
                    \Else
                        \State $b \gets b'$
                    \EndIf
                \EndIf
            \EndFor
        \EndFor
    \EndFor
\EndWhile
\EndFunction
\end{algorithmic}

\subsection{A Simple Worked Example}
\label{SimpleWorkedExample}

One can give a simple worked example to show how BOSS works for a very small case and why the two-step procedure (Algorithm \ref{TWO_STEP}) is needed when BOSS does not give a minimal permutation yet are unable to make any improvements in their final permutations. The example has four nodes and four edges.

Let the true model be this:

\begin{verbatim}
Graph Nodes:
X1;X2;X3;X4

Graph Edges:
1. X1 --> X2
2. X1 --> X3
3. X2 --> X4
4. X3 --> X4
\end{verbatim}

\noindent These are all of the 24 possible permutations and the edge counts of their implied DAGs. The minimal models are marked with '*'. The score here is edge count. A parent X of Y is included in the model, by Pearl, just in case $Y \perp\!\!\!\!/\!\!\!\!\!\perp X | Prefix(Y) \setminus {X}$, where $Prefix(Y)$ is the set of nodes prior to Y in the permutation. We mark as ``TRUTH'' permutations for DAGs in the Markov equivalence class of the true DAG.

\begin{verbatim}
permutation = [X1, X2, X3, X4] edges = 4 TRUTH A
permutation = [X1, X2, X4, X3] edges = 6
permutation = [X1, X3, X2, X4] edges = 4
permutation = [X1, X3, X4, X2] edges = 6
permutation = [X1, X4, X2, X3] edges = 6
permutation = [X1, X4, X3, X2] edges = 6
permutation = [X2, X1, X3, X4] edges = 4 TRUTH B
permutation = [X2, X1, X4, X3] edges = 6
permutation = [X2, X3, X1, X4] edges = 5
permutation = [X2, X3, X4, X1] edges = 5 
permutation = [X2, X4, X1, X3] edges = 6 
permutation = [X2, X4, X3, X1] edges = 5 
permutation = [X3, X1, X2, X4] edges = 4 TRUTH C
permutation = [X3, X1, X4, X2] edges = 6
permutation = [X3, X2, X1, X4] edges = 5
permutation = [X3, X2, X4, X1] edges = 5 
permutation = [X3, X4, X1, X2] edges = 6
permutation = [X3, X4, X2, X1] edges = 5 
permutation = [X4, X1, X2, X3] edges = 6 
permutation = [X4, X1, X3, X2] edges = 6
permutation = [X4, X2, X1, X3] edges = 6 
permutation = [X4, X2, X3, X1] edges = 5 
permutation = [X4, X3, X1, X2] edges = 6 
permutation = [X4, X3, X2, X1] edges = 5 
\end{verbatim}

Consider starting from this permutation: 

\begin{verbatim}
START: permutation = [X4, X2, X3, X1] edges = 5 
\end{verbatim}

\noindent It is possible to get back to a minimal model by moving one node? No. Below are all the one-variable moves that reduce the score (rearranging the above), for BOSS. 

\bigskip
\noindent Moving X1 will not reduce the score:
    
\begin{verbatim}
permutation = [X1, X4, X2, X3] edges = 6
permutation = [X4, X1, X2, X3] edges = 6 
permutation = [X4, X2, X1, X3] edges = 6 
permutation = [X4, X2, X3, X1] edges = 5 
\end{verbatim}

\noindent Nor X2:

\begin{verbatim}
permutation = [X2, X4, X3, X1] edges = 5 
permutation = [X4, X2, X3, X1] edges = 5 
permutation = [X4, X3, X2, X1] edges = 5 
permutation = [X4, X3, X1, X2] edges = 6 
\end{verbatim}

\noindent Nor X3:

\begin{verbatim}
permutation = [X3, X4, X2, X1] edges = 5 
permutation = [X4, X3, X2, X1] edges = 5 
permutation = [X4, X2, X3, X1] edges = 5 
permutation = [X4, X2, X1, X3] edges = 6 
\end{verbatim}

\noindent Not X4 either: 

\begin{verbatim}
permutation = [X4, X2, X3, X1] edges = 5 
permutation = [X2, X4, X3, X1] edges = 5 
permutation = [X2, X3, X4, X1] edges = 5 
permutation = [X2, X3, X1, X4] edges = 5
\end{verbatim}

\noindent These moves at least yield the \textit{same} minimal score = 5:

\begin{verbatim}
X2:

(1) permutation = [X2, X4, X3, X1] edges = 5  

X3:

(2) permutation = [X3, X4, X2, X1] edges = 5 

X4:

(3) permutation = [X2, X4, X3, X1] edges = 5 
(4) permutation = [X2, X3, X4, X1] edges = 5 
(5) permutation = [X2, X3, X1, X4] edges = 5
\end{verbatim}

\noindent These are the permutations with \textit{lower} scores:

\begin{verbatim}
(a) permutation = [X1, X2, X3, X4] edges = 4
(b) permutation = [X2, X1, X3, X4] edges = 4
(c) permutation = [X3, X1, X2, X4] edges = 4
\end{verbatim}

\noindent One could get from (3) to (a) or (b) in one move by moving X1. For example,		

\begin{verbatim}
START: permutation = [X4, X2, X3, X1] edges = 5 
\end{verbatim}

\noindent then

\begin{verbatim}
(3) permutation = [X2, X3, X4, X1] edges = 5 
\end{verbatim}

\noindent then

\begin{verbatim}
(a) permutation = [X1, X2, X3, X4] edges = 4
\end{verbatim}

So it's possible to get from START to TRUTH A in two moves, though the first must be a ``sideways'' (i.e., non-greedy) move that doesn't lower the score. What's needed here is a move that orients an unshielded collider by moving two nodes simultaneously--that is, a ``two step'' move.

\section{Evaluation}

All comparisons are done in the TETRAD freeware using the \textit{algcomparison} tool (\cite{ramsey2020algcomparison}),

\subsection{Oracle Performance}

By adjusting the calculation of the iterated Grow-Shrink algorithm, it is possible to run BOSS given an independence test rather than a score, as in Solus et al. Also, it's possible to run BOSS from a d-separation oracle along the same lines, where d-separation facts are obtained (faithfully) from an examination of the true graph. Solus et al. did not do this; their study rather approximated oracle performance in their Figure 1 and Figure 2 by thresholding partial correlations calculated using the true covariance matrix. However, a d-separation oracle study is feasible; if one uses d-separation as an oracle for BOSS using the Tetrad implementation, one gets the following, for their simulation parameters--i.e, 10 nodes with average degrees 1 through 9:

\begin{verbatim}
  Alg  avgDegree    AP    AR   AHP    AR    SHD     E
    1       1.00  1.00  1.00  1.00  1.00      -  0.01
    1       2.00  1.00  1.00  1.00  1.00      -  0.05
    1       3.00  1.00  1.00  1.00  1.00      -  0.07
    1       4.00  1.00  1.00  1.00  1.00      -  0.10
    1       5.00  1.00  1.00  1.00  1.00      -  0.14
    1       6.00  1.00  1.00     *  1.00      -  0.20
    1       7.00  1.00  1.00  1.00  1.00      -  0.12
    1       8.00  0.93  1.00     *  1.00  32.00  0.10
    1       9.00  1.00  1.00     *  1.00      -  0.01
\end{verbatim}

\noindent For each average degree, a new random graph was generated and then BOSS was applied using a d-separation oracle, and results tabulated.\footnote{The fact that statistics are very slightly less than 1 for some of the graphs suggests suggests possibly an intermittent bug. So far the identity of this bug has not been discovered.} 
 
\subsection{Unfaithful Examples from Raskutti and Uhler and Solus et al.}
\label{UnfaithfulExamples}

One thing that can be said about the relationship between GSP (either TSP or ESP), SP, and BOSS is that the counterexamples given in Raskutti and Uhler and Solus et al. are not problematic for the BOSS algorithms, so judging from these examples alone, there is no reason to distinguish the Boss algorithms from SP. These examples are to illustrate that while certain dependencies exist between the various assumptions studied in those algorithms, the implications are strong in the case that counterexamples exist in for their converses. These counterexamples are all unproblematic for BOSS; each gives the same model as SP, for BOSS, provided one finishes each with the two-step procedure. Included in the comparison are the simple 4-node path canceling algorithm mentioned in the introduction, plus the counterexample from Theorem 2.4 from Raskutti and Uhler, plus four counterexamples from Solus et al. In each case, an exhaustive list of conditional independencies is provided; these lists are unfaithful, so the question is how well the correct graph can be reconstructed from the unfaithful independencies.

There's a very simple type of comment one can make of these examples and any others of the same sort. BOSS treats these examples in the manner of a greedy selection game. The goal is to start with one permutation and find another permutation that has a score that's less than the score of the given permutation. The algorithms included in the comparison are BOSS, introduced above, and the sparsest permutation algorithm (SP). The question is whether any of these unfaithful cases cause the procedures to get stuck--that is, where there is a non-correct permutation, all of the adjacent permutations of which (according to the game) have higher scores.

Running the models, under these conditions, one gets the following counts for numbers of unique CPGDAGs found in a random simulation study, for each of our models, where each algorithm is run 500 times over and all of the unique CPDAGs for DAGs generated by the algorithms are collected up. Each example as it is given in the original articles is as a list of conditional independence facts, together with a true DAG. For each of the algorithms the number of unique CPDAGs found across all runs is listed

\begin{enumerate}
\item Simple 4-node path canceling model that GES should get right.

\begin{verbatim}
1 _||_ 3 | 2
2 _||_ 4 | 1, 3
1 _||_ 4

BOSS = 1, SP = 1
\end{verbatim}

\item Raskutti and Uhler's Theorem 2.4 SMR $\not\Rightarrow$ Restricted Faithfulness.

\begin{verbatim}
1 _||_ 3 | 2
2 _||_ 4 | 1, 3
1 _||_ 2 | 4

BOSS = 1, SP = 1
\end{verbatim}

\item Solus Theorem 11, TSP $\not\Rightarrow$ Faithfulness counterexample (Figure 6).

\begin{verbatim}
1 _||_ 5 | 2, 3
2 _||_ 4 | 1, 3
3 _||_ 5 | 1, 2, 4
1 _||_ 4 | 2, 3, 5
1 _||_ 4 | 2, 3

BOSS = 1, SP = 1
\end{verbatim}

\item Solus Theorem 12, ESP $\not\Rightarrow$ TSP (Figure 7).

\begin{verbatim}
1 _||_ 2 | 4
1 _||_ 3 | 2
2 _||_ 4 | 1, 3

BOSS = 1, SP = 1
\end{verbatim}

\item Solus Theorem 11, SMR $\not\Rightarrow$ ESP (Figure 8).

\begin{verbatim}
1 _||_ 3 | 2
2 _||_ 4 | 1, 3
4 _||_ 5

BOSS = 1, SP = 1
\end{verbatim}

\item Solus Theorem 12, TSP $\not\Rightarrow$ Orientation Faithfulness (Figure 11).

\begin{verbatim}
1 _||_ 3
1 _||_ 5 | 2, 3, 4
4 _||_ 6 | 1, 2, 3, 5
1 _||_ 3 | 2, 4, 5, 6

BOSS = 1, SP = 1
\end{verbatim}
\end{enumerate}

\noindent In call cases, the models found by the two algorithms are unique, and the BOSS result is identical to the result SP gives. This is consistent with the conjecture above that BOSS is correct under the SMR assumption.

Parenthetically, the model (3) (Solus et al.'s Figure 6) is interesting in that all algorithms (including SP) agree on an 8-edge model, though this is because Algorithm \ref{dag} is being used to build the DAGs, which uses Algorithm \ref{GrowShrink} to build parent sets. If in place of Algorithm \ref{GrowShrink} Pearl's method had been used instead for building DAGs, as Solus et al. and Raskutti and Uhler do, each algorithm would have yielded a graph with 7 edges, as reported in Solus et al. Figure 6. So the two procedures, as noted above, are not equivalent. Here, $1 \perp\!\!\!\perp 4 | 2, 3$ is being treated as a path cancellation. In practice, with large models, under brute faithfulness, if the Verma-Pearl method is used, BOSS returns graphs of about the same quality but much more slowly for linear, Gaussian models. At least, that is, up to a certain average degree which is found heuristically.

\subsection{Comparison to Lu et al.}

Another recent algorithm that reports accuracies for linear, Gaussian Bayes net search in the same general range is the Triplet $A^*$ algorithm (\cite{lu2021improving}). Theoretically, Triplet $A^*$ looks to correct errors in unshielded colliders, as above, but does not finish by applying a set of implied orientation rules such as the Meek rules (\cite{meek2013causal}), so it doesn't guarantee that the output is a CPDAG. Also, it resolves conflict of orientations between triplets arbitrarily. Nevertheless, the results are quite good in terms of accuracy statistics.\footnote{This point is due to Wayne Lam.} There seems to be a relationship between Triplet $A^*$, SP and BOSS, in that all of these algorithms make comparison, at least by implication, of scores across all (or most) variable permutations in a certain class. In the case of Triplet $A^*$, the permutations are over subsets of variables used to determine unshielded collider orientations. For SP and BOSS, the permutations considered are for the entire dataset. The idea of comparing all or most permutations over certain variables classes seems like promising; perhaps other such strategies could be devised. In any case, among these three algorithms, BOSS has the advantage of speed; it is not fast, but it can scale comfortably to hundreds of variables in reasonable time in simulation for sparse models, without losing accuracy.\footnote{Note that for large sparse problems one doesn't lose accuracy if one proceeds by only moving variables to the left in a permutation order, instead of moving each variable to every position in the permutation at each step (left or right). This saves considerable time for large problem, though the results are not reported here. Also, with large problems, one has the option of running these on a supercomputer where more memory is available to store cached scores, thus speeding up the procedure. As it stands, all simulations in this paper were done on a MacBook Pro with an Apple M1 processor and 16 G RAM; for the larger problems caching is turned off.} So there is an advantage in elapsed time.

Notably, Lu et al. (\cite{lu2021improving}) do a comparison of GSP to Triplet $A^*$, with Triplet $A^*$ comparing favorably to GSP. In a future draft, we will repeat some of that work here, but for now the reader is referred to the Lu et al. paper for this comparison.

Here is data for BOSS corresponding to Figure 3 in Lu et al. Here, as in Lu et al. for this figure, the number of variables is fixed at 20, with an averages degree of 4, with coefficients drawn from $(0.2, 0.8)$. Here and in subsequent sections, except where explicitly noted, graphs are ``random forward''--that is to say, Erdos-Renyi undirected graphs directed as DAGs. (A simulation using scale-free graphs is given below, the exception.) These figures may be compared directly to their Figure 3. The statistics reported are as follows:

\begin{verbatim}
avgDegree = Extra column for avgDegree
sampleSize = Extra column for sampleSize
AP = Adjacency Precision
AR = Adjacency Recall
AHP = Arrowhead precision
AHR = Arrowhead recall
SHD = Structural Hamming Distance
E = Elapsed Time in Seconds
\end{verbatim}

\noindent Note that graphs are being compared to the true CPDAG. This data:

\begin{verbatim}
  sampleSize    AP    AR   AHP   AHR    SHD     E
       50.00  0.88  0.59  0.78  0.49  40.20  0.19
      100.00  0.93  0.72  0.80  0.59  31.00  0.08
      200.00  0.96  0.80  0.71  0.59  28.40  0.13
      500.00  0.99  0.97  0.98  0.97   2.40  0.14
     1000.00  0.95  0.99  0.84  0.89  11.00  0.16
     2000.00  0.99  0.99  0.99  0.98   1.40  0.17
     5000.00  0.99  1.00  0.96  0.96   3.00  0.21
    10000.00  1.00  1.00  1.00  1.00      -  0.22
    20000.00  0.98  0.98  0.96  0.94   4.20  0.26
    50000.00  0.97  0.99  0.93  0.96   4.60  0.27
   100000.00  0.99  1.00  1.00  0.96   1.60  0.28
   200000.00  0.99  1.00  1.00  0.96   1.80  0.48
\end{verbatim}

\noindent Although the comparison is not completely fair since fewer runs were done in the table above, AP and AR here are about the same as precision and recall reported in Figure 3 for Triplet $A^*$, the implication being that since Triplet $A^*$ dominates GSP, BOSS does as well.

There has been some discussion or coefficient ranges for searches of this general sort. These may be expanded to for ranges that do not exclude an interval about zero and for which the range is wider. With coefficients in $(-4, 4)$:

\begin{verbatim}
  sampleSize    AP    AR   AHP   AHR    SHD     E
       50.00  0.72  0.66  0.49  0.48  53.80  0.30
      100.00  0.89  0.89  0.87  0.76  21.40  0.18
      200.00  0.91  0.88  0.84  0.79  20.80  0.14
      500.00  0.90  0.93  0.88  0.72  20.60  0.19
     1000.00  0.97  0.93  0.91  0.89   9.80  0.19
     2000.00  0.92  0.95  0.83  0.90  14.60  0.24
     5000.00  0.87  0.92  0.77  0.79  21.80  0.25
    10000.00  1.00  0.98  1.00  0.97   1.60  0.21
    20000.00  0.96  0.97  0.95  0.95   6.00  0.20
    50000.00  0.99  0.99  0.99  0.99   1.00  0.36
   100000.00  1.00  1.00  1.00  1.00      -  0.35
   200000.00  1.00  0.99  1.00  0.99   0.40  0.51
\end{verbatim}

\noindent With coefficients in $(-10, 10)$:

\begin{verbatim}
  sampleSize    AP    AR   AHP   AHR    SHD     E
       50.00  0.78  0.76  0.58  0.68  40.00  0.36
      100.00  0.89  0.87  0.88  0.86  19.00  0.18
      200.00  0.85  0.86  0.78  0.81  25.60  0.18
      500.00  0.95  0.95  0.93  0.93   8.60  0.19
     1000.00  0.92  0.92  0.91  0.72  18.80  0.17
     2000.00  0.98  0.98  0.97  0.98   3.60  0.17
     5000.00  0.92  0.96  0.86  0.91  12.80  0.26
    10000.00  0.97  0.99  0.95  0.95   4.60  0.25
    20000.00  0.99  0.99  0.99  0.99   1.20  0.27
    50000.00  1.00  0.99  1.00  0.99   0.60  0.34
   100000.00  0.92  0.98  0.88  0.92  13.40  0.40
   200000.00  1.00  1.00  1.00  1.00      -  0.58
\end{verbatim}

\noindent Parenthetically, the width of coefficient ranges needs to be more restrictive for denser graphs for good performance; performance is given here only for sparse graphs with average degree 4, as noted above.

Next, data is simulated to correspond to Figure 6 in \cite{lu2021improving}. In that figure, sample size is fixed at 500, and the number of variables is fixed at 60, coefficients drawn from $0.2, 0.8$, with the average degree of the graph increasing (judging from the plot) from about 2 to about 5 in increments of 0.5. This range of average degrees is increased to a range from 2 to 12 below. FGES here is an optimized implementation of GES (\cite{ramsey2017million}). CPC is the Conservative PC algorithm \cite{ramsey2012adjacency}.

\begin{verbatim}
Algorithms:

1. BOSS using SEM BIC Score
2. CPC using Fisher Z test
3. FGES using SEM BIC Score

Graphs are being compared to the True CPDAG.

AVERAGE STATISTICS

All edges

  Alg  avgDegree  sampleSize    AP    AR   AHP   AHR     SHD      E
    1       2.00      500.00  0.97  0.99  0.85  0.95   10.80   2.05
    2       2.00      500.00  1.00  0.94  0.64  0.63   28.60   0.11
    3       2.00      500.00  0.98  0.97  0.81  0.91   13.80   0.13
    1       4.00      500.00  0.98  0.98  0.94  0.92   16.60   5.58
    2       4.00      500.00  0.99  0.71  0.64  0.49  101.60   0.10
    3       4.00      500.00  0.91  0.94  0.80  0.85   51.00   0.35
    1       6.00      500.00  0.99  0.99  0.97  0.99   11.60   8.80
    2       6.00      500.00  0.98  0.60  0.61  0.40  186.80   0.14
    3       6.00      500.00  0.70  0.88  0.51  0.68  223.60   1.38
    1       8.00      500.00  0.99  0.98  0.96  0.98   21.00  12.78
    2       8.00      500.00  0.91  0.43  0.55  0.27  329.80   0.18
    3       8.00      500.00  0.62  0.85  0.43  0.60  386.80   3.60
    1      10.00      500.00  0.99  0.99  0.98  0.98   17.60  27.76
    2      10.00      500.00  0.88  0.31  0.53  0.19  475.20   0.19
    3      10.00      500.00  0.50  0.73  0.30  0.44  684.80   7.71
    1      12.00      500.00  0.98  0.97  0.97  0.96   39.80  51.93
    2      12.00      500.00  0.89  0.26  0.56  0.17  587.20   0.21
    3      12.00      500.00  0.49  0.69  0.29  0.41  835.40   9.93
\end{verbatim}

\noindent Again, the comparison is not completely fair here, since only one run was done, but BOSS  for average degrees 2 does better than Triplet $A^*$ on every statistic except AHP, and for average degree 4 does better on every statistic commonly reported. Moreover, BOSS is feasible and does well out to an average degree of 12 here. (For average degree 14, not reported here, statistics become much worse, for reasons not completely understood.) For CPC (as with PC) there are good adjacency precision out to an average degree of 10, though with adjacency recall falling off, and orientation statistics falling to chance or below rapidly. For FGES, adjacency precision falls to chance by average degree 10, faster than adjacency recall, with orientation precision falling to chance by average degree 8.

Note that while the comparison above to Lu et al.'s Figure 3 does not give a compelling reason to prefer BOSS to Triplet $A^*$, the comparison to their Figure 6 does, both in terms of accuracy and in terms of elapsed time, BOSS is able to complete a problem with 60 nodes with an average degree of 12 in 51.93 seconds; Triplet $A^*$ takes far longer (5 hours, personal communication) to complete a problem with 60 nodes and an average degree of 5.

\subsection{Comparison on Smaller Sample Sizes}

Simulations may be run following the pattern in a well-known paper by Nandy et al. [\cite{nandy2018high}), Figure 5 (top row). There, performances are given for four algorithms as ROC plots, plotting true positive rate (i.e., recall). For each, four combinations of simulation parameters are studied, as given in the table below (Table 3 in \cite{nandy2018high}). For GES, the parameter varied is $\lambda_n = c(ln(N))$ for a linear, Gaussian BIC score; in TETRAD, $c$ is called the ``penalty discount'', which will be varied. For the first parameter combination with N = 50, the first two choices for penalty discount throw singularity exceptions, so these are not shown. Otherwise, these simulations are all within the (easy) purview of BOSS. The Table below gives true positive rate and false positive rate for penalty discount ranging from 1.0 to 5.0 in increments of 0.5, along with several other statistics. These will be rendered as ROC plots, though the ATPR and AFPR rates for adjacencies in the table below may be compared to Figure 5 in \cite{nandy2018high}. In general, the performances have been lifted considerably. For these, coefficients were chosen uniformly from U(-1, 1),\footnote{The coefficient range in \cite{nandy2018high} was not clear, so a generic such range was chosen that does not exclude an interval about zero.} error variances from (1, 3), and used BOSS with the BIC score, without using the two-step procedure. This is for just one run.

\begin{verbatim}
100 nodes, 100 edges, N = 50

  Alg  penaltyDiscount    AP    AR   AHP   AHR  ATPR     AFPR     SHD      E
    1             2.00  0.43  0.62  0.26  0.49  0.62     0.02  244.00  13.01
    2             2.50  0.62  0.58  0.40  0.45  0.58  7.28E-3  156.00   6.71
    3             3.00  0.90  0.56  0.67  0.42  0.56  1.22E-3   99.00   6.63
    4             3.50  0.94  0.49  0.71  0.36  0.49  6.12E-4  103.00   3.81
    5             4.00  0.96  0.43  0.66  0.28  0.43  4.09E-4  114.00   6.55
    6             4.50  0.98  0.40  0.80  0.23  0.40  2.04E-4  116.00   5.77
    7             5.00  0.98  0.40  0.80  0.23  0.40  2.04E-4  116.00   2.21
    
100 nodes, 200 edges, N = 100

  Alg  penaltyDiscount    AP    AR   AHP   AHR  ATPR     AFPR     SHD      E
    1             1.00  0.36  0.77  0.26  0.60  0.77     0.05  672.00  60.48
    2             1.50  0.69  0.76  0.63  0.70  0.76     0.01  243.00  21.07
    3             2.00  0.87  0.72  0.82  0.64  0.72  4.47E-3  164.00   8.06
    4             2.50  0.93  0.70  0.91  0.61  0.70  2.04E-3  150.00  13.78
    5             3.00  0.96  0.67  0.92  0.55  0.67  1.23E-3  160.00  14.31
    6             3.50  0.97  0.61  0.91  0.52  0.61  8.20E-4  175.00  12.87
    7             4.00  0.99  0.59  0.94  0.48  0.59  2.05E-4  178.00   8.28
    8             4.50  0.98  0.59  0.95  0.49  0.59  4.11E-4  180.00   7.16
    9             5.00  0.99  0.56  0.97  0.46  0.56  2.06E-4  189.00   7.33  
    
100 nodes, 300 edges, N = 150

  Alg  penaltyDiscount    AP    AR   AHP   AHR  ATPR     AFPR     SHD      E
    1             1.00  0.58  0.79  0.54  0.75  0.79     0.03  489.00  81.33
    2             1.50  0.83  0.76  0.82  0.75  0.76  9.34E-3  235.00  55.56
    3             2.00  0.93  0.76  0.91  0.72  0.76  3.27E-3  185.00  36.50
    4             2.50  0.97  0.73  0.96  0.70  0.73  1.44E-3  181.00  24.69
    5             3.00  0.95  0.70  0.93  0.65  0.70  2.05E-3  212.00  38.11
    6             3.50  0.97  0.66  0.95  0.62  0.66  1.44E-3  228.00  34.13
    7             4.00  0.93  0.61  0.88  0.56  0.61  2.68E-3  272.00  25.71
    8             4.50  0.95  0.60  0.90  0.55  0.60  2.07E-3  272.00  13.80
    9             5.00  0.99  0.61  0.98  0.58  0.61  4.14E-4  243.00  21.67
    
100 nodes, 400 edges, N = 200

  Alg  penaltyDiscount    AP    AR   AHP   AHR  ATPR     AFPR     SHD       E
    1             1.00  0.72  0.84  0.69  0.81  0.84     0.03  396.00  118.95
    2             1.50  0.90  0.81  0.88  0.79  0.81  7.54E-3  236.00   75.42
    3             2.00  0.96  0.78  0.95  0.75  0.78  2.46E-3  210.00   40.58
    4             2.50  0.98  0.76  0.95  0.74  0.76  1.44E-3  214.00   45.15
    5             3.00  0.99  0.75  0.96  0.72  0.75  8.24E-4  217.00   50.43
    6             3.50  0.99  0.72  0.96  0.69  0.72  4.13E-4  240.00   41.94
    7             4.00  0.99  0.69  0.95  0.64  0.69  4.14E-4  269.00   30.02
    8             4.50  1.00  0.66  0.96  0.62  0.66  2.08E-4  286.00   19.65
    9             5.00  0.99  0.61  0.96  0.58  0.61  4.17E-4  321.00   21.92
\end{verbatim}

In a future draft, these will be plotted--that is, ATPR will be plotted against AFPR for ease of comparison, but a comparison shows that all of these curves are lifted with respect to Figure 5 (top row) of Nandy et al.

Unfortunately, Nandy et al.'s larger simulations in their Figure 3 lie outside the easy purview of BOSS using linear BIC, so that comparison will not be given. One run was done for each statistic; as time permits, the simulation will be re-run for multiple runs with statistics averaged.

\subsection{Comparison with Large Numbers of Variables}

There is another dimension for comparison, where the average degree of the graph is fixed to something fairly sparse and the number of variable is increased instead. In the following table the average degree is fixed at 4 and the number of variables ranges from 50 up to 300. Because score caching requires a fair amount of memory for large problems, it is turned off for so that these simulation can be run on a laptop. Also, the sample size for these is fixed at N = 1000; for smaller sample sizes (say, N = 500 or fewer) adjacency precision can suffer.

\begin{verbatim}
  Alg  numMeasures    AP    AR   AHP    AR    SHD        E
    1        50.00  1.00  1.00  1.00  1.00      -    17.40
    1       100.00  1.00  0.99  0.99  0.99   3.00   147.73
    1       150.00  0.98  1.00  0.98  1.00  11.00   596.24
    1       200.00  0.99  1.00  0.99  1.00   6.00  1433.32
    1       250.00  0.99  1.00  0.99  1.00   9.00  2535.05
    1       300.00  0.97  1.00  0.97  1.00  40.00  6059.28      
\end{verbatim}

\noindent While estimation accuracy is good for all of these runs, they can become quite slow. The slowest, here, for 300 variables, is 1.6 hours.

\subsection{Comparisons on Mixed and Discrete Datasets}

For data with mixtures of continuous and discrete columns, the Conditional Gaussian BIC score by Andrews et al. (\cite{andrews2018scoring}) can be used for both FGES and BOSS. Using the method proposed by Lee and Hastie (\cite{lee2013structure}), 200 or 1000 records of data are simulated with 25 continuous variables and 25 discrete, 3 categories per variable for the discrete variables, with linear connection functions among the linear variables. The variables are randomly ordered. The Lee and Hastie method treats discrete values as ordinal. The Conditional Gaussian method has an option to discretize continuous variables that are children of discrete variables; not using this option is more accurate, and it is not onerous in our simulations. Here we use $score_{edge}$ and set the penalty discount for the conditional Gaussian score to 2.

\begin{verbatim}
Algorithms:

1. FGES using Conditional Gaussian BIC Score
2. BOSS using Conditional Gaussian BIC Score

Graphs are being compared to the True CPDAG.

AVERAGE STATISTICS

All edges

  Alg  avgDegree  sampleSize    AP    AR   AHP   AHR     SHD      E
    1       2.00      200.00  1.00  0.74  0.78  0.21   41.00   1.62
    2       2.00      200.00  1.00  0.74  0.80  0.24   40.00   3.31
    1       2.00     1000.00  1.00  0.96  0.61  0.71   19.00   3.98
    2       2.00     1000.00  0.98  0.94  0.86  0.51   22.00   9.79
    1       4.00      200.00  0.92  0.56  0.72  0.27  115.00   3.22
    2       4.00      200.00  0.97  0.61  0.95  0.42   92.00   3.99
    1       4.00     1000.00  0.99  0.76  0.74  0.63   69.00   6.08
    2       4.00     1000.00  0.99  0.75  0.87  0.59   65.00  20.38
    1       6.00      200.00  0.94  0.49  0.70  0.36  176.00   1.66
    2       6.00      200.00  0.91  0.51  0.73  0.33  179.00   5.00
    1       6.00     1000.00  0.90  0.63  0.75  0.53  146.00   7.81
    2       6.00     1000.00  0.93  0.67  0.80  0.60  127.00  28.80
    1       8.00      200.00  0.67  0.34  0.43  0.19  360.00   4.17
    2       8.00      200.00  0.94  0.50  0.93  0.46  219.00   9.51
    1       8.00     1000.00  0.87  0.59  0.80  0.53  211.00  27.23
    2       8.00     1000.00  0.95  0.61  0.92  0.57  176.00  71.88
\end{verbatim}

\noindent The preliminary conclusion (based on just one run) is that BOSS has a consistent, though not huge, advantage over FGES with the same conditional Gaussian score, with the algrotihm configured as above. The advantage is more evident with denser graphs.

Discrete data may also be simulated using the Lee and Hastie method. Again, data for 50 variables are simulated with sample sizes 200 and 1000, with 3 categories per variable. Here, the BDeu score (\cite{heckerman1995learning}) is used. Again, only one run was done for this table.

\begin{verbatim}
Algorithms:

1. FGES using BDeu Score
2. BOSS using BDeu Score

Graphs are being compared to the True CPDAG.

AVERAGE STATISTICS

All edges

  Alg  avgDegree  sampleSize    AP    AR   AHP   AHR     SHD     E
    1       2.00      200.00  1.00  0.88  0.79  0.54   25.00  0.41
    2       2.00      200.00  1.00  0.90  0.84  0.60   19.00  1.06
    1       2.00     1000.00  1.00  0.98  0.73  0.84   14.00  0.18
    2       2.00     1000.00  1.00  1.00  0.69  0.82   14.00  1.21
    1       4.00      200.00  1.00  0.56  0.95  0.20  118.00  0.29
    2       4.00      200.00  1.00  0.62  0.86  0.35   98.00  1.80
    1       4.00     1000.00  0.98  0.81  0.68  0.61   67.00  0.43
    2       4.00     1000.00  0.99  0.83  0.77  0.68   52.00  1.95
    1       6.00      200.00  1.00  0.41  1.00  0.20  202.00  0.28
    2       6.00      200.00  0.99  0.46  0.82  0.32  181.00  1.54
    1       6.00     1000.00  0.98  0.58  0.84  0.44  151.00  0.50
    2       6.00     1000.00  0.97  0.63  0.80  0.50  139.00  3.71
    1       8.00      200.00  1.00  0.34  0.84  0.22  281.00  0.28
    2       8.00      200.00  1.00  0.39  0.83  0.29  262.00  2.03
    1       8.00     1000.00  0.98  0.54  0.78  0.42  212.00  0.51
    2       8.00     1000.00  0.99  0.56  0.83  0.47  195.00  2.54 
\end{verbatim}

\noindent For this type of data and in these simulation ranges, there appears to be a small though consistent advantage for BOSS in the SHD statistic, but more runs needs to be done. In any case, because the advantage is so small, there is little reason to prefer BOSS in this condition to FGES, although there seems to be no disadvantage.

\subsection{Scale-free Graphs}

In previous comparisons, directed Erdos-Renyi graphs have been used; here is a brief comparison using scale-free graphs, which are in some contexts more realistic. These graphs are generated using the method of \cite{bollobas2003directed}, with alpha = 0.41, beta = 0.54, delta-in = 0.2, and delta-out = 0.1, the authors defaults (except for delta-out, which has been moved slightly away from zero). A linear, Gaussian SEM simulation was used. Here BOSS uses $score_{edge}$.

\begin{verbatim}
1. FGES using SEM BIC Score
2. BOSS using SEM BIC Score

Graphs are being compared to the True CPDAG.

AVERAGE STATISTICS

All edges

  Alg  sampleSize  numMeasures  EdgesT  EdgesEst    AP    AR   AHP   AHR     SHD      E
    1      500.00        10.00    9.00      8.10  0.95  0.84  0.99  0.84    3.20   0.03
    2      500.00        10.00    9.00      7.60  0.98  0.82  0.94  0.78    3.60   0.04
    1     1000.00        10.00   13.00     12.90  0.90  0.88  0.86  0.79    7.20   0.02
    2     1000.00        10.00   13.00     13.00  0.86  0.85  0.91  0.62    9.80   0.02
    1    10000.00        10.00   11.00     11.20  0.93  0.94  0.78  0.72    5.60   0.02
    2    10000.00        10.00   11.00     10.60  0.98  0.95  1.00  0.84    2.20   0.02
    1      500.00        20.00   23.00     24.40  0.81  0.79  0.75  0.69   23.20   0.03
    2      500.00        20.00   23.00     19.50  0.90  0.76  0.81  0.61   16.80   0.11
    1     1000.00        20.00   25.00     24.80  0.87  0.85  0.88  0.78   15.30   0.03
    2     1000.00        20.00   25.00     22.40  0.93  0.83  0.95  0.71   13.20   0.10
    1    10000.00        20.00   25.00     25.00  0.97  0.97  0.98  0.97    2.70   0.03
    2    10000.00        20.00   25.00     24.40  0.99  0.96  0.98  0.94    2.70   0.14
    1      500.00        50.00   81.00    103.90  0.62  0.75  0.57  0.65  130.80   0.69
    2      500.00        50.00   81.00     72.70  0.79  0.70  0.65  0.54   89.30   3.51
    1     1000.00        50.00   60.00     64.20  0.83  0.88  0.84  0.86   34.30   0.06
    2     1000.00        50.00   60.00     58.90  0.87  0.85  0.85  0.73   36.20   2.47
    1    10000.00        50.00   78.00    133.40  0.72  0.94  0.69  0.84  135.40  23.10
    2    10000.00        50.00   78.00     74.50  0.98  0.93  0.96  0.87   17.20  11.07
\end{verbatim}

BOSS has an advantage here in terms of SHD, though improvements could be made. If larger sample sizes are available, these are helpful for the larger models.

\section{Conclusion}

Some novelties were introduced for permutation DAG search. First, it is noted that assuming brute faithfulness, Markov blankets can be recovered of variables with respect to sets containing those variables even when path cancellation (or the independence equivalent to path cancellation) obtains, when building DAGs using Algorithm \ref{GrowShrink}. This is something that any permutation algorithm could take advantage of. Second, a novel (albeit simple) method is introduced for traversing the space of permutations that's not depth-first as in other algorithms, but rather aims to reverse incorrect orientations about a node. It is shown that this traversal, in combination with the assumption of brute faithfulness, can improve accuracy of search. In fact, one may conjecture, based on performance on counterexamples in Raskutti and Uhler (\cite{raskutti2018learning}) and Solus et al. (\cite{solus2017consistency}) that such a search will always return the same graph as SP if the SMR assumption of Raskutti and Uhler holds. Some simulation examples are included for small and large N for the linear, Gaussian case; in all cases BOSS has good lift against existing results. For mixed continuous/discrete data, there is some lift for denser models as compared to GES (FGES), though for discrete models BOSS and GES perform about the same, each using the same score across algorithms.

Future work includes exploring (as for instance in \cite{bernstein2020ordering}) the use of this kind of BOSS permutation search for latent variables models. It's possible that the increased accuracy will be of some benefit there for the linear, Gaussian case. One possibility is to simply replace FGES by BOSS in the GFCI algorithm (\cite{ogarrio2016hybrid}) for increased accuracy. It would also be helpful to adapt the method to other variable types and function types where path cancellation is a topic of interest, such as the linear, non-Gaussian case. It is not clear \textit{a priori} that there would be any advantage, so these cases need to be explored explicitly. 

In terms of the draft, an empirical example was not given; one needs to be added. Tables need to be formatted for LaTeX and figures rendered with some of the information in them. Most simulations need to be redone averaging statistics over more runs, now that the forms of the comparisons are worked out. Also, some explicit comparison may be added of BOSS to GSP, over and above what was provided by Lu et al.

\bibliographystyle{abbrv}
\bibliography{bibliography}

\section{Appendix}

\subsection{Details of Implementation}

A scorer is coded in Java along the lines of Teyssier and Kohler (\cite{teyssier2012ordering}) in an object-oriented fashion. It is initialized by scoring a particular permutation, and methods are made available to move a variable to the right or to the left in permutation $O$ by one index, or to move it to a new index, or to swap the positions of two variables. Also, a bookmarking facility is provided, where the state of the scorer at any point in time can be saved and returned to later. Scoring is done using Algorithm \ref{GrowShrink}, returning alternatively a BIC score or an edge count. Algorithm \ref{BOSS} is implemented in an optimized way using the Teyssier and Kohler scorer, using move operations with bookmarking. Another method is made available in the scorer to return the Markov blanket of a node as calculated using Algorithm \ref{GrowShrink}. Algorithm \ref{TWO_STEP} is implemented in an optimized way by making use of the swap method and the Markov blanket method; the only graph checking it does is to check triangles, which can be done by looking at the parent sets of variables, already calculated in the scorer. The BOSS algorithm is then coded, as given in Algorithms \ref{BOSS}. Scores may alternatively be cached or not. The Java implementation of this algorithm will be made available in the TETRAD freeware; the code will be publicly available in the repository for the TETRAD project at https://github.com/cmu-phil/tetrad.\footnote{One doesn't expect difficulty coding this algorithm in other languages; it makes very few demands on the language. For the Teyssier et al. scorer, object orientation would seem to be preferable.} In the TETRAD project, datasets may be simulated \textit{ad libitum}; nevertheless, the datasets used to generate the tables in the evaluation sections will be made available for comparison to other methods.

\end{document}